\documentclass[10pt]{article}

\usepackage[letterpaper,margin=1in,columnsep=0.25in]{geometry}

\usepackage{graphicx}
\usepackage{booktabs}
\usepackage{amsmath,amssymb,amsfonts}
\usepackage{multirow}
\usepackage{xcolor}
\usepackage{xspace}
\usepackage{microtype}
\usepackage{enumitem}
\usepackage{makecell}
\newcommand{\Name}{W2S\xspace}
\newcommand{\IRName}{Skill-IR\xspace}
\newcommand{\DataSetName}{WSASkill\xspace}
\usepackage{natbib}
\usepackage[
    colorlinks=true,
    linkcolor=blue,
    citecolor=blue,
    urlcolor=blue
]{hyperref}
\usepackage{subcaption}

\makeatletter
\DeclareRobustCommand\onedot{\futurelet\@let@token\@onedot}
\def\@onedot{\ifx\@let@token.\else.\null\fi\xspace}
 
\def\ie{\emph{i.e}\onedot}

\makeatother

\title{Workflow-to-Skill: Skill Creation via Routing-Workflow-Semantics-Attachments Decomposition}

\author{Yuyang Zhang$^{1}$ \quad Xinyuan Han$^{2}$ \quad Xudong Jiang$^{1}$ \quad Run Wang$^{1}$\\
$^{1}$Key Laboratory of Aerospace Information Security and Trusted Computing, \\Ministry of Education, School of Cyber Science and Engineering, Wuhan University \\
$^{2}$Nanchang University\\
}

\date{}

\begin{document}

\twocolumn[{
\begin{@twocolumnfalse}
\maketitle

\begin{abstract}
Large language model agents increasingly rely on \emph{Skills} to encode procedural knowledge, including when to invoke a capability, how to decompose a task, what constraints to follow, and how to verify intermediate outcomes. 
Despite their importance, high-quality Skills are still largely hand-written, making them difficult to scale across domains, tools, and execution environments. This paper studies how to automatically construct executable Skills from heterogeneous interaction evidence, such as demonstrations, agent trajectories, tool-use traces, and execution logs. We argue that this is not a standard summarization problem: historical traces are often fragmented across scenarios, contain redundant or accidental steps, and may omit low-frequency but safety-critical operations. To address this challenge, we introduce \IRName, an intermediate representation that interprets a Skill from a workflow perspective and decomposes its content into three complementary components: Workflow structure, execution Semantics, and runtime Attachments. Together, these WSA components capture the structural, behavioral, and operational elements required for executable Skills, including task decomposition, step-level execution requirements, control-flow conditions, verification procedures, and safety-critical state management. Building on \IRName, we propose \Name, a trace-to-skill construction framework that converts historical execution evidence into reusable agent Skills. \Name segments traces into procedural units, induces local Skill drafts from individual paths, aligns and merges shared structures across scenarios, reconciles conditional branches, and compresses redundant steps while preserving verification, approval, rollback, and state-management behaviors with evidence and confidence annotations. Experiments on 70 skills show that \Name improves behavioral replay consistency over summarization- and prompting-based baselines, with an improvement of 10.5\%. 
These results suggest that reliable Skill generation requires treating historical traces as evidence for executable runtime specifications rather than as text to be compressed.
\end{abstract}

\vspace{1em}
\end{@twocolumnfalse}
}]

\section{Introduction}

LLM agents~\citep{luo2025large} are rapidly evolving from systems that merely generate responses into runtime systems that execute workflows~\citep{wang2024agent}, invoke tools~\citep{shi2025tool}, and read and write state~\citep{xie2024osworld}. As agents assume these runtime responsibilities, an abstraction is needed to package reusable agent capabilities and specify how they should be applied across tasks. In this transition, skills have emerged as a key abstraction for organizing such reusable capabilities~\citep{ling2026agent}. A skill specifies what capability an agent can reuse, together with clear instructions about when it should be activated and how it should be used~\citep{li2026skillsbench}. In this sense, a skill is not merely a prompt fragment; it is a runtime specification intended to guide agent behavior across future tasks~\citep{xu2026agent}. Recent practice suggests that skills can improve the reliability, transferability, and maintainability of agent behavior, and that prompts and tool-use procedures are increasingly being reorganized or re-implemented as skills~\citep{jiang2026sok,ling2026agent}. Because skills can provide a scalable interface for accumulating, transferring, and operationalizing agent experience, they are likely to remain an important building block in future agent systems and have been attracting growing attention from both industry and academia~\citep{zhou2026comprehensive}.

However, despite their demonstrated value, current skills are largely manually authored, making them costly to scale and difficult to keep aligned with evolving usage scenarios, tool environments, and execution requirements~\citep{liu2026skillforge,ma2026skillgen}.
Fortunately, rich evidence of task-oriented behavior already exists in the form of interaction traces, tool calls, expert demonstrations, user feedback, and execution logs, providing a natural basis for automated skill induction~\citep{huang2026raw}.
Yet current approaches to skill generation are often limited to summarizing traces, rather than producing structured runtime specifications that future agents can reliably reuse~\citep{li2026skilltracer,yang2026survey}.
As a result, the induced skills may overfit incidental details, omit critical preconditions or recovery procedures, and become difficult to verify or maintain.
A more principled formulation is therefore needed: skill induction should transform execution data into structured, reusable procedural knowledge that can guide future agent behavior.

Skill creation differs from ordinary summarization in its objective. 
Summarization typically compresses historical content according to semantic salience~\citep{tang2023understanding}, whereas skill creation aims to reconstruct procedural knowledge that can support future execution~\citep{wu2026agent}. 
For an induced skill to be reusable, it must preserve not only the main intent of prior traces, but also the runtime structure that determines how an agent should act: triggering conditions, task decomposition ~\citep{yao2022react}, tool-use policies, constraints, failure handling, and validation criteria~\cite{shinn2023reflexion}. These elements are often functionally distinct rather than hierarchically important, and therefore may be merged or omitted by a purely summary-oriented process. Thus, skill creation requires transforming traces into a compact but structured workflow, rather than simply producing a concise description of what happened~\citep{zhou2026skillgenbench}.

Our key insight is that the proper unit of skill induction should not be a textual instruction, but a structured runtime specification, especially for automated skill generation. 
Unlike ordinary text summarization, which compresses source data into salient semantic content\citep{radford2021learning}, skill creation must preserve the operational properties that make a skill executable and reusable. 
A generated skill should not merely describe what the source data is about; it should specify when the skill applies, how the task proceeds, how local decisions are made, and what runtime safeguards constrain execution.

\begin{figure*}[t]
  \centering

  \begin{subfigure}[t]{0.32\linewidth}
    \centering
    \includegraphics[height=0.22\textheight,keepaspectratio]{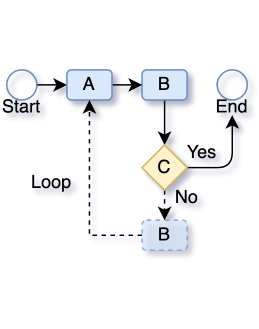}
    \caption{Workflow Backbone \(W\)}
    \label{fig:skill-ir-w}
  \end{subfigure}
  \hfill
  \begin{subfigure}[t]{0.32\linewidth}
    \centering
    \includegraphics[height=0.22\textheight,keepaspectratio]{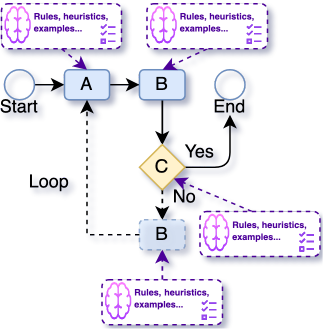}
    \caption{Operational Semantics \(S\)}
    \label{fig:skill-ir-s}
  \end{subfigure}
  \hfill
  \begin{subfigure}[t]{0.32\linewidth}
    \centering
    \includegraphics[height=0.22\textheight,keepaspectratio]{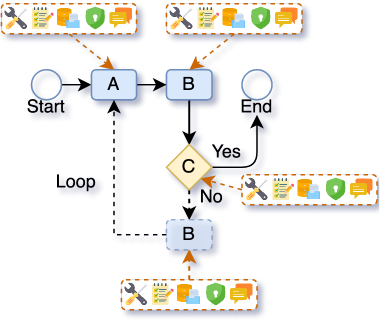}
    \caption{Runtime Attachments \(A\)}
    \label{fig:skill-ir-a}
  \end{subfigure}

  \caption{Overview of the \IRName{} representation. \IRName{} models a skill as a structured runtime specification composed of three complementary components: the Workflow Backbone \(W\), Operational Semantics \(S\), and Runtime Attachments \(A\). Together, these components separate workflow structure, node-level behavior, and runtime constraints, enabling skills to be reconstructed and evaluated as executable specifications rather than free-form summaries.}
  \label{fig:Skill-IR}
\end{figure*}

To this end, we introduce \IRName, an intermediate representation that converts interaction traces into computable objects before rendering them into executable agent instructions. 
\IRName represents a skill with a routing header and three runtime components. 
The \emph{routing header}, comprising the front matter and description used for skill discovery, specifies when a skill should be considered applicable. As shown in Figure~\ref{fig:Skill-IR}, the \emph{workflow backbone} captures the control structure of execution, including workflow nodes and directed transitions among them (when the skill has at least two workflow nodes connected by a directed edge). 
\emph{Node-level semantics} define the local objectives and decision criteria that govern branch, retry, and termination behavior recorded in the workflow paths.
\emph{Runtime attachments} describe the operational context required by execution, such as tools, scripts, resources, references, templates, configuration constraints, and output requirements. 
Together with the routing header, this decomposition separates when a skill applies, how it executes, how decisions are made, and how runtime effects are constrained.

Building on this insight, we propose \Name, an evidence-driven skill construction framework. 
\Name aligns traces into scenarios, extracts path-level observations, and generates grounded skill drafts for each path. 
It then fuses shared workflow nodes, reconciles branches and conflicts, and compresses redundancy while preserving critical. 
The drafted intermediate representation is finally rendered as a reusable agent skill.

Experiments on multi-scenario agent traces show that \Name improves replay-based behavioral fidelity compared with summarization- and prompting-based baselines.  
These results suggest that historical traces should be treated as execution evidence for inducing runtime specifications, rather than as documents to be compressed. 
More broadly, they show that reliable agent learning from experience benefits from explicit intermediate representations, path-level evidence tracking, and runtime-aware structure.

Our contributions are threefold:
\begin{itemize}[leftmargin=*, itemsep=0pt, topsep=0pt]
    \item We identify automated agent skill generation as a structured induction task rather than a trace summarization of past data. To support this formulation, we introduce \IRName, which represents skill through a routing header, a workflow backbone, node-level semantics, and runtime attachments.
    \item We propose \Name, an evidence-driven framework that constructs \IRName from path-level execution traces by aligning scenarios, merging compatible patterns, and preserving execution-critical constraints.
    \item Experiments show that, under the same interaction evidence, \Name consistently outperforms Anthropic Skill Creator in both structural fidelity and behavioral consistency on \DataSetName dataset.
\end{itemize}

\section{Related Work}

\subsection{Agent Skills}

LLM-based agents are increasingly moving from monolithic prompting toward modular procedural abstractions~\citep{shi2025tool,ruan2023tptu}. 
An \emph{agent skill}, which is intended to persist across related tasks and sessions, is a reusable operational package that specifies when it should be activated, how the agent should proceed, and what task-specific resources, scripts, tools, or constraints should be used during execution~\citep{ling2026agent}. 
Unlike a low-level tool or API, it does not necessarily expand the primitive action space; rather, it organizes existing instructions, actions, and resources into a repeatable procedure. 
In this sense, skills serve as a procedural layer between high-level task conditioning and concrete environment interaction: memory stores facts or preferences, tools expose primitive capabilities, whereas skills describe how those capabilities should be composed for recurring tasks~\citep{jiang2026sok,zhou2026comprehensive}.

Recent agent runtimes and skill formats make this abstraction explicit. A skill is typically packaged with metadata for discovery, natural-language instructions for execution, and optional auxiliary files such as scripts, references, templates, or examples~\citep{hermes2026skills}. 
Such designs also adopt progressive disclosure that agents first inspect lightweight descriptions to decide whether a skill is relevant, and load the full procedural content only when needed. 
This makes skills attractive for long-horizon agent systems, where reusable procedures must be invoked selectively without flooding the context with all available experience. 
As a result, skills are becoming a central form of \emph{procedural memory} for LLM agents: they are editable, versionable, portable across compatible runtimes, and auditable as explicit artifacts rather than latent model behavior~\citep{jiang2026sok,wu2026agent}.

\subsection{Trace-grounded Skill Induction}

A growing line of work studies how such skills can be acquired automatically from agent experience~\citep{wang2024agent,xia2026skillrl,wang2026skillx,huang2026raw}. 
We refer to this direction as \emph{trace-grounded skill induction}: the process of converting historical interaction or execution traces into reusable skill artifacts. 
A trace may contain user requests, observations, intermediate reasoning, tool calls, environment actions, execution outcomes, corrections, and repeated failure or success patterns~\citep{ni2026trace2skill}. 
The key idea is to treat these traces not as passive logs to be retrieved or summarized, but as behavioral evidence from which future operating procedures can be reconstructed~\citep{li2026arise}.

Existing methods instantiate this idea in different forms. 
Agent Workflow Memory induces reusable workflows from past web-agent trajectories and retrieves them to guide future action generation~\citep{wang2024agent}. 
Agent Skill Induction further represents induced skills as executable programs, enabling the system to verify skill correctness through execution rather than using free-form textual lessons alone~\citep{wang2025inducing}. 
AutoSkill abstracts recurring user requirements and interaction patterns into explicit, maintainable skills that can be updated and injected across sessions~\citep{yang2026autoskill}. 
SkillRL distills raw trajectories into a hierarchical skill library and lets the skill library co-evolve with the agent policy during reinforcement learning~\citep{xia2026skillrl}. 
Trace2Skill analyzes multiple executions in parallel, extracts trajectory-local lessons, and consolidates them into transferable skill directories that can either deepen existing human-written skills or create new ones from scratch~\citep{ni2026trace2skill}. 
Together, these studies show that experience can be compressed into persistent procedural artifacts, improving agent success, efficiency, transfer, and long-term adaptation without retraining the underlying model.

However, this line of work also exposes a fundamental representation challenge~\citep{liu2026skillsvote}.
The usefulness of a skill does not depend only on whether it preserves salient task content, but also on whether it preserves the runtime structure that makes the behavior executable~\citep{liang2026skill}. 
If traces are compressed into free-form summaries or loosely organized lessons, the resulting skill may lose operational details such as activation conditions, workflow stages, branch criteria, retry and fallback rules, tool-use requirements, validation checks, and termination conditions~\citep{li2026arise}. 
These details are not cosmetic: changing them can alter whether a skill is invoked, which path the agent follows, when it retries or stops, and what output constraints are enforced~\citep{liang2026skill}. 
Thus, effective trace-grounded skill induction requires more than extracting important observations from past traces. 
It must reconstruct the structured runtime specification that governs future agent behavior.

Our work builds on this perspective. 
Rather than treating trace-derived skills as generic summaries of past experience, we study them as structured operational artifacts whose control logic, node-level execution semantics, and runtime attachments jointly determine how an agent acts. 
This framing allows us to analyze trace-grounded skill induction as a conversion from interaction evidence to executable specifications.

\section{WSA Representation}
While summarization primarily produces a compact natural-language description, we view a skill as a structured object with explicitly defined inputs, outputs, and execution semantics.
Specifically, we model a skill as the combination of a routing header and a runtime specification.
The routing header determines when the skill should be selected (\textit{i.e.}, description), while the runtime specification determines how the skill behaves once selected. 
We decompose the skill into a routing header $R$ and a runtime specification:
\begin{equation}
  \mathrm{Skill} = (R,\; \underbrace{W + S + A}_{\text{runtime specification}}),
\end{equation}
where $R$ determines when the skill should be selected,
while $W$, $S$, and $A$ jointly determine how it behaves once selected.
\(W\) is the \textit{workflow backbone} that captures the execution skeleton and node dependencies; \(S\) is the \textit{operational semantics} that specify how nodes are interpreted and executed; and \(A\) denotes \textit{runtime attachments} that define the operational boundary, including tools, resources, constraints, state operations, validation checks, and output schemas.

\subsection{Workflow Backbone}

The workflow backbone \(W\) captures the execution structure of a skill:
\begin{equation}
  W = (N,E),
\end{equation}
where \(N\) is a set of workflow nodes and \(E\) is a set of directed links between nodes.
Each node represents an abstract execution unit, and each directed link represents an ordering or dependency relation between execution units.
We say a skill \emph{has a workflow backbone} when \(|N| \geq 2\) and the nodes are connected by at least one directed link, indicating a multi-step execution flow.
A single-node skill (\(|N| = 1\)) has no structural backbone to recover, even though it still possesses operational semantics and possibly runtime attachments.
Thus, \(W\) describes the skeleton of the procedure: what units exist and how execution may move among them.

\subsection{Operational Semantics}

Operational semantics \(S\) specify how the workflow nodes in \(W\) should be interpreted and executed.
For each relevant node, \(S\) records its behavioral meaning, including its local objective, decision logic, execution conditions, and criteria for success or failure.
In this sense, \(S\) complements the workflow backbone by assigning executable meaning to the structural units.

\subsection{Runtime Attachments}

Runtime attachments \(A\) specify the external and contextual dependencies required by skill execution.
They include the resources, interfaces, constraints, state interactions, validation requirements, and output commitments that define the operational boundary of the skill.
Unlike \(W\), which captures execution structure, and \(S\), which captures node-level behavior, \(A\) captures what the skill depends on or constrains at runtime.

\begin{table*}[t]
\centering
\small
\resizebox{\linewidth}{!}{%
\begin{tabular}{cllll}
\toprule
 & Components & Type & Representative Use Case & Typical skill \\
\midrule
T0 & none & Prompt Fragment & ``Write in a professional tone'' for email drafting. &  shakespearean-english \\
T1 & A & Attachment Wrapper & Expose a code formatter tool or API documentation template. & weather-api \\
T2 & S & Semantic Guideline & ``Prioritize user safety over feature completeness'' in content moderation. & code-review-checklist \\
T3 & S+A & Semantic Resource & Use a customer database under the rule ``only query active accounts''. &  react-best-practices\\
T4 & W & Bare Workflow & ``Fetch data → Parse → Return result'' for simple ETL tasks. &  prototype\\
T5 & W+A & Tool-Driven Workflow & ``Check inventory → Reserve stock → Send confirmation email'' in order fulfillment. & azure-compliance \\
T6 & W+S & Semantically Guided Workflow & ``If urgent, escalate; else queue'' in customer support routing. & to-issues \\
T7 & W+S+A & Full Runtime Workflow & ``Retrieve user profile → Apply business rules → Call payment API → Log transaction'' in checkout. &  test-driven-development\\
\bottomrule
\end{tabular}%
}
\caption{Skill types defined by the presence of workflow backbone \(W\),
operational semantics \(S\), and runtime attachments \(A\). A skill is assigned \(W\) when \(|N| \geq 2\) with at least one directed edge;
single-node skills (\(|N| = 1\)) are treated as having no \(W\).}
\label{tab:taxonomy}
\end{table*}

\subsection{Skill Type Coverage}

The \IRName model induces eight skill types, summarized in
Table~\ref{tab:taxonomy}. These types define the coverage target for the dataset. 
Non-workflow skills (\(|N| \leq 1\)) are included because they test whether a SkillCreator can reconstruct semantic or resource-oriented capabilities without a multi-step backbone to anchor on.
Workflow-oriented skills are especially important because they test whether the model can reconstruct persistent execution contracts.

Our dataset construction aims to cover this taxonomy explicitly. For each skill, annotators or automated analyzers assign a WSA profile, identify the dominant skill type, and record which components are observable from the available traces.
This makes it possible to report reconstruction performance not only in aggregate, but also by skill type.

\begin{table}[t]
\centering
\small
\begin{tabular}{ll}
\toprule
Statistic & Content \\
\midrule
Reference skills & 70 skills\\
WSA skill-type taxonomy & 8 skill types \\
\bottomrule
\end{tabular}
\caption{Current dataset snapshot used by the \Name reconstruction
study. We count a skill scenario type as a distinct WSA path or use case,
rather than as a raw dialogue turn.}
\label{tab:dataset-statistics}
\end{table}

\section{Dataset Construction}

We construct \DataSetName to evaluate whether a method can reconstruct executable agent skills from interaction evidence. The dataset contains 70 reference skills and covers the eight WSA skill types defined in Table~\ref{tab:taxonomy}. Each reference skill is first analyzed under the \IRName framework: we extract its workflow backbone \(W\), operational semantics \(S\), and runtime attachments \(A\). This produces a structured reference representation that specifies both the organization of the skill and the runtime behavior it is expected to preserve.

For workflow-bearing skills, we parse the workflow backbone and enumerate W-paths from the parsed graph. Each W-path corresponds to one INPUT-to-OUTPUT route, with loops collapsed into loop units and branch or loop annotations retained. This path-level enumeration turns a skill into a set of controlled runtime scenarios, allowing the dataset to cover not only the main execution flow but also alternative branches, loop behavior, fallback cases, and termination outcomes.

We then collect interaction evidence for each W-path. Specifically, for every enumerated path, we collect 10 traces or real interaction instances that demonstrate the expected behavior of that path. These traces serve as the input evidence for skill reconstruction, while the extracted WSA representation provides the reference used for replay-based behavioral evaluation.

This construction makes \DataSetName a behavior-centered benchmark rather than a collection of raw dialogue logs. By combining WSA component extraction, deterministic W-path enumeration, and multiple traces per path, the dataset supports fine-grained evaluation across different skill structures and runtime scenarios.

\section{Methodology}
\Name is a \IRName-guided trace-grounded skill induction method. The goal is not to produce a fluent summary of demonstrations, but to recover a reusable runtime specification whose workflow, node-level rules, and runtime bindings are grounded in interaction evidence. 
We use notation to describe the interface between stages, not to claim a closed-form estimator.
Given trace evidence \(\mathcal{D}\), \Name first reconstructs an intermediate WSA representation and then renders that representation as a reusable skill:
\begin{equation}
  \mathcal{D}
  \xrightarrow{\mathrm{WSA parse}}
  \widehat{\Theta}
  =
  (\widehat{W},\widehat{S},\widehat{A})
  \xrightarrow{\mathrm{skill generation}}
  \widehat{\mathcal{S}} .
  \label{eq:method-interface}
\end{equation}
The central methodological claim is that reconstruction should be organized
around WSA component recovery. Without this structure, a model may produce a
plausible instruction file while losing branch coverage, decision criteria,
tool constraints, or validation behavior.

\begin{figure*}[htbp]
  \centering
  \includegraphics[width=0.85\linewidth]{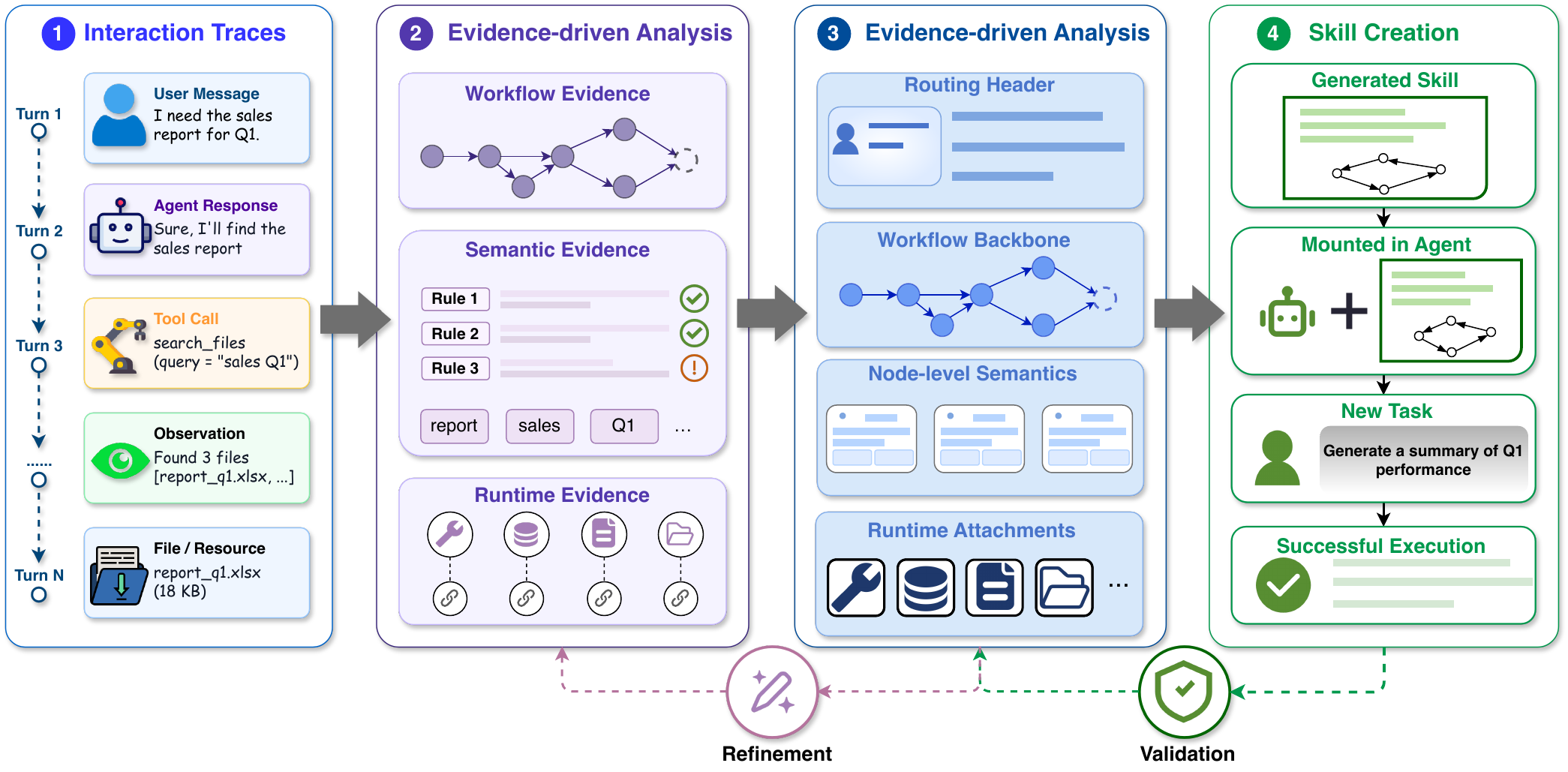}
  \caption{Overview of our \Name. The framework converts historical agent interaction traces into structured evidence, including workflow evidence, semantic evidence, and runtime evidence. 
These evidence types are integrated into an intermediate skill representation with four components, \ie, a routing header, a workflow backbone, node-level semantics, and runtime attachments. 
Finally, the intermediate representation is rendered as a reusable executable skill and refined through validation feedback.}
  \label{fig:method}
\end{figure*}

\subsection{WSA Reconstruction from Traces}

\Name begins by converting interaction traces into WSA-oriented
evidence. This step does not assign numerical support scores. Instead, it
separates the observations needed to reconstruct a runtime skill:
\begin{equation}
  \mathcal{E}
  =
  \{\mathcal{E}_W,\mathcal{E}_S,\mathcal{E}_A\},
  \label{eq:evidence-interface}
\end{equation}
where \(\mathcal{E}_W\) contains evidence about execution units and ordering,
\(\mathcal{E}_S\) contains evidence about node functions and decision rules,
and \(\mathcal{E}_A\) contains evidence about tools, resources, validation
checks, state requirements, and output contracts. The abstraction also records
provenance: whether a statement is directly observed in a trace, inferred
from repeated behavior, or unobserved. This provenance discipline is central
to the method because the output skill is a persistent runtime contract, not a
summary of examples.

The WSA reconstruction stage treats a trace as a multi-signal object. User
requests reveal activation conditions and task scope; agent actions reveal
workflow order; justifications reveal decision criteria; failed or rejected
paths reveal node-local constraints; and final responses reveal output
contracts. \Name keeps these signals separated so that formatting
conventions are not mistaken for workflow steps and isolated decisions are
not promoted into global rules.

The workflow component is reconstructed as
\(\widehat{W}=(\widehat{N},\widehat{E})\).
When the reference skill contains only a single execution unit (\(|N| = 1\)), the WSA reconstruction stage produces a degenerate \(\widehat{W}\) with one node and no edges, and the generation step skips workflow instructions.
Here \(\widehat{N}\) denotes
execution units such as intake, retrieval, filtering, planning, checking, and
response generation, while \(\widehat{E}\) denotes directed links among those
units. The method deliberately keeps \(W\) structural: branching, looping,
approval, fallback, and termination are not encoded as separate elements of
\(W\). They are node functions recovered in \(S\).

The semantic and attachment components are then aligned to the recovered
workflow. Symbolically, the semantic layer is a node-indexed interpretation:
\begin{equation}
  \widehat{S}: \widehat{N} \rightarrow
  \{\mathrm{goal},\mathrm{criteria},\mathrm{outcome},\mathrm{quality}\}.
  \label{eq:semantic-interface}
\end{equation}
This notation is intentionally schematic: it states the scope of semantic
reconstruction rather than a fixed schema. It allows the method to attach
goals, decision criteria, validation rules, examples, and outcomes to the node
where they affect behavior. Runtime attachments are similarly scoped:
\begin{equation}
  \widehat{A}
  =
  \widehat{A}_{\mathrm{global}}
  \cup
  \widehat{A}_{\mathrm{node}} .
  \label{eq:attachment-interface}
\end{equation}
Global attachments define the skill-wide environment, while node-local
attachments define resources, actions, checks, or output obligations that
apply only at a particular execution unit. This joint WSA reconstruction makes
errors inspectable: a failure can be attributed to a missing node, a wrong
link, a wrong node function, or a mis-scoped runtime attachment.

\subsection{WSA-Constrained Skill Generation}

Once WSA components are identified, \Name generates a skill document
under explicit structural constraints. The reconstructed skill must include
activation conditions, a coherent workflow, node-level semantics for
decision-bearing nodes, runtime attachments, validation requirements, and
output behavior. The generation step follows three rules.

First, every major instruction must be traceable to W, S, or A evidence.
Second, branch alternatives and rare paths should remain explicit rather than
being absorbed into generic prose. Third, uncertainty must be represented as
uncertainty. If the traces do not demonstrate a tool argument, failure mode,
or branch condition, the reconstructed skill may mark it as unobserved or
request clarification, but should not fabricate a precise rule.

This constrained synthesis reduces two common reconstruction errors. The
first is \emph{over-compression}, where distinct branches or validation cases
are collapsed into a single broad instruction. The second is
\emph{over-generalization}, where the model invents workflow steps,
decision criteria, or attachments that were not supported by the traces.
Path-aware WSA generation keeps rare behavior visible, while evidence
discipline limits unsupported generalization.

Operationally, generation aims to produce a skill text whose parsed WSA
structure matches the recovered components:
\begin{equation*}
  \mathrm{Parse}_{WSA}(\widehat{\mathcal{S}})
  =
  (\widehat{W},\widehat{S},\widehat{A})
  \approx
  \left(
    \widehat{W}(\mathcal{D}),
    \widehat{S}(\mathcal{D}),
    \widehat{A}(\mathcal{D})
  \right).
  \label{eq:wsa-parse-match}
\end{equation*}
For non-workflow skills (\(|N| = 1\)), \(\widehat{W}\) is a degenerate single-node graph with no edges.
The approximation is deliberate: the final skill must be readable and usable, but it should not erase component boundaries. Workflow instructions state the node-link structure; semantic instructions explain node functions and criteria; attachment instructions bind resources, tools, checks, and output requirements to the appropriate scope.

The generation stage is therefore constrained by three writing obligations.
First, workflow content must preserve the recovered nodes and links. Second,
semantic content must remain node-local when it governs a specific decision,
validation, fallback, or termination behavior. Third, attachment content must
state the scope of resources and checks. These obligations are qualitative
constraints, not claims that the generator solves a particular optimization
problem.

\subsection{Feedback Refinement}

\Name includes a feedback loop that evaluates and revises the draft
before producing the final skill. The loop has three checks. The
\emph{coverage check} compares the reconstructed WSA components against the
trace evidence and flags missing path segments, branch outcomes, decision
criteria, or attachments. The \emph{consistency check} detects contradictions, such as a workflow that requires approval but an attachment section that makes the corresponding action unconditional. 
The \emph{executability check} reviews whether the reconstructed skill is actionable for a downstream agent: steps must be ordered, decision criteria must be placed at the right nodes, and validation or output requirements must be specific enough to guide behavior.

When a check fails, \Name returns targeted feedback to the generation step. 
The feedback is WSA-local: a missing edge is repaired in \(W\), a vague criterion is repaired in \(S\), and a missing tool or validation binding is repaired in \(A\).
The revised skill is then re-parsed into \((\widehat{W},\widehat{S},\widehat{A})\) and checked again. This iterative process continues until no high-priority WSA error remains or a fixed refinement budget is reached. 
In this way, feedback refinement is not a generic polishing step; it is a structured attempt to improve reconstruction fidelity along the same components used for evaluation.

We write the refinement loop as:
\begin{equation}
  \widehat{\mathcal{S}}^{(t+1)}
  =
  \mathrm{Revise}
    \left(
    \widehat{\mathcal{S}}^{(t)},
    \Delta_W^{(t)},
    \Delta_S^{(t)},
    \Delta_A^{(t)}
  \right).
  \label{eq:feedback-interface}
\end{equation}
The feedback terms are obtained by comparing the parsed candidate skill with the WSA evidence:
\begin{equation}
  (\Delta_W^{(t)},\Delta_S^{(t)},\Delta_A^{(t)})
  =
  \mathrm{Check}
  \left(
    \mathrm{Parse}_{WSA}(\widehat{\mathcal{S}}^{(t)}),
    \mathcal{E}
  \right).
  \label{eq:feedback-check}
\end{equation}
\(\Delta_W\) contains node-link coverage errors, \(\Delta_S\) contains node-function or criterion errors, and \(\Delta_A\) contains missing or mis-scoped runtime bindings. 
The process stops when no high-priority WSA error remains or the refinement budget is exhausted.

The repair operation is typed. A W-level repair may split a collapsed node, restore a missing link, or separate two alternative paths. 
An S-level repair may rewrite a vague instruction into a node-local predicate, add a missing accept/reject criterion, or move a termination condition from the backbone description into the semantic record of the relevant node. 
An A-level repair may bind a resource to the node that uses it, mark an approval requirement as mandatory, or restore an output schema omitted by the previous draft. 
Since the feedback is expressed in the same WSA language as the evaluation, the method aligns its revisions with the fidelity dimensions later reported by the benchmark.

\section{Experiments}
We conduct experiments to evaluate whether our framework can faithfully reconstruct skills from workflow evidence. Our evaluation focuses on replay-based behavioral fidelity. We begin by introducing the experimental setup, metrics, and then report the main empirical results.

\subsection{Experimental Setup}

\textbf{Dataset.} We evaluate our method on \DataSetName, a benchmark organized around the WSA skill taxonomy.
For each reference skill, \DataSetName provides the corresponding WSA profile, workflow backbone when applicable, and replay scenarios used to assess behavioral fidelity. 
These scenarios instantiate different execution paths, decision outcomes, and attachment-use cases, enabling us to compare reconstructed skills against their reference behavior under controlled replay settings.

\noindent\textbf{Baseline.} We use Anthropic Skill Creator (ASC) as the baseline. ASC is an official skill-construction workflow proposed by Anthropic that converts interaction evidence into a reusable \texttt{SKILL.md} file through a structured interview and drafting process. It is directly comparable to our setting because both methods aim to construct reusable agent skills from prior interaction evidence.

\noindent\textbf{Metrics}. We evaluate reconstructed skills using \emph{replay-based behavioral fidelity}. For each WSA type, we replay the same evaluation scenarios with the reconstructed skill and compare its behavior against the reference skill. The score measures whether the reconstruction preserves the expected execution behavior, including task progression, decision outcomes, tool-use constraints, validation requirements, and final responses. Higher scores indicate stronger behavioral consistency with the reference skill.

\subsection{Main Results}

Table~\ref{tab:main-results} reports the replay-based behavioral fidelity results across WSA skill types. Overall, \Name outperforms Anthropic Skill Creator (ASC) on most categories, achieving an average score of 0.503 compared with 0.455 for ASC. This indicates that the WSA-guided reconstruction more effectively preserves the expected execution behavior of reference skills.

The main exception is T5, where ASC achieves a higher score than \Name. T5 represents skills that combine a workflow backbone with runtime attachments but do not include explicit operational semantics. In this setting, the model must recover how tools, resources, and constraints should be bound to workflow steps without clear semantic guidance. This makes the generated skill sensitive to attachment placement and tool-binding details, which can lead to behavior mismatches during replay. We view this as evidence that attachment-heavy workflows require stronger mechanisms for scoping resources and constraints in future versions of \Name.

\begin{table}[t]
\centering
\small
\begin{tabular}{lrrrr}
\toprule
Type & \Name & ASC & Gap \\
\midrule
T0 & \textbf{0.623} & 0.520 & +0.103\\
T1 & \textbf{0.522} & 0.407 & +0.115\\
T2 & \textbf{0.760} & 0.638 & +0.122 \\
T3 & \textbf{0.573} & 0.533 & +0.040\\
T4 & \textbf{0.276} & 0.227 & +0.049\\
T5 & 0.480 & \textbf{0.550} & -0.070 \\
T6 & \textbf{0.515} & 0.450 & +0.065 \\
T7 & \textbf{0.652} & 0.613 & +0.039 \\
\midrule
Average & \textbf{0.503} & 0.455 & +0.048 \\
\bottomrule
\end{tabular}
\caption{Replay-based behavioral fidelity results by WSA skill type. Gap is computed as \Name minus ASC; higher scores indicate stronger behavioral consistency with the reference skill.}
\label{tab:main-results}
\end{table}

These results suggest that WSA-guided skill reconstruction generally improves behavioral preservation over ASC, while also revealing that attachment-heavy workflow skills remain the most challenging category for the current method.

\section{Conclusion}

We present automated agent skill generation as a structured induction problem rather than a trace summarization problem. 
Our key idea is that a reusable skill should preserve the runtime information needed for future execution, including the workflow structure, node-level decision semantics, and operational constraints.
To this end, we introduce \IRName, a structured representation that makes these components explicit and separates executable skill reconstruction from ordinary text compression.
Building on \IRName, our framework reconstructs skills from behavioral evidence and evaluates them by whether they preserve the runtime contract of reference skills. Our evaluation moves beyond surface-level text similarity and measures whether generated skills preserve the runtime contract of reference skills.

\section*{Open Science}
In accordance with open science policies, this paper adheres to principles that promote transparency, accessibility, and reproducibility.
All source code, and supplementary materials are publicly available at \href{https://github.com/Q1ngSong/RWSA}{https://github.com/Q1ngSong/RWSA}, enabling verification, reuse, and further exploration of our methods.

\bibliographystyle{unsrtnat}
\bibliography{references}

\end{document}